\documentclass{article}
\usepackage{kr}

\usepackage{times}
\usepackage{soul}
\usepackage{url}
\usepackage[hidelinks]{hyperref}
\usepackage[utf8]{inputenc}
\usepackage[small]{caption}
\usepackage{graphicx}
\usepackage{amsmath}
\usepackage{amsthm}
\usepackage{booktabs}
\usepackage{tabularx}
\usepackage{algorithm}
\usepackage{algorithmic}
\usepackage{xcolor}

\newtheorem{definition}{Definition}

\usepackage{amsthm}

\theoremstyle{remark}
\newtheorem{remark}{Remark}

\usepackage{listings}

\lstdefinelanguage{Prolog}{
	morekeywords={:-, ?, is, not, fail, true, listing, and, if},
	sensitive=true,
	morecomment=[l]{\%},
	morestring=[b]",
}

\lstdefinestyle{pddl}{
	language=sh,                
	basicstyle=\ttfamily\footnotesize, 
	keywordstyle=\bfseries\color{blue}, 
	stringstyle=\color{red},     
	commentstyle=\color{green},  
	showstringspaces=false,      
	frame=none,               
	breaklines=true,            
	breakautoindent=true,       
	escapeinside={(*@}{@*)},    
}

\newenvironment{italicquote}
  {\begin{quote}\itshape}
  {\end{quote}}

\title{Towards a Common Framework for Autoformalization}

\author{%
Agnieszka Mensfelt$^*$\and
David Tena Cucala$^*$\and
Santiago Franco$^*$\and \\
Angeliki Koutsoukou-Argyraki${^{*,\dag}}$\and 
Vince Trencsenyi$^*$\and
Kostas Stathis$^*$
\affiliations
$^*$Department of Computer Science, Royal Holloway, University of London, \\ Egham Hill, Egham, TW200EX, UK\\
$^\dag$Department of Computer Science and Technology, University of Cambridge, \\ William Gates Building, 15 JJ Thomson Avenue, Cambridge, CB3 0FD, UK\\
\emails
\{agnieszka.mensfelt, david.tenacucala, santiago.francoaixela, angeliki.koutsoukouargyraki, 
vince.trencsenyi, kostas.stathis\}@rhul.ac.uk
}

\begin{document}

\maketitle

\begin{abstract}
\textit{Autoformalization} has emerged as a term referring to the automation of formalization—specifically, the formalization of mathematics using interactive theorem provers (proof assistants). Its rapid development has been driven by progress in deep learning, especially large language models (LLMs). More recently, the term has expanded beyond mathematics to describe the broader task of translating informal input into formal logical representations. At the same time, a growing body of research explores using LLMs to translate informal language into formal representations for reasoning, planning, and knowledge representation—often without explicitly referring to this process as autoformalization. As a result, despite addressing similar tasks, the largely independent development of these research areas has limited opportunities for shared methodologies, benchmarks, and theoretical frameworks that could accelerate progress. The goal of this paper is to review—explicit or implicit—instances of what can be considered autoformalization and to propose a unified framework, encouraging cross-pollination between different fields to advance the development of next generation AI systems. 
\end{abstract}

\section{Introduction} 

Drawing on the long tradition of formalizing human reasoning through mathematics and logic, AI pioneers~\cite{mccarthy1959,kowalski1974,newell1976} incorporated ideas from formal systems  into functional and logical
programming languages for formalizing knowledge and solving practical problems. Building on these foundations, researchers in natural language processing~\cite{weizenbaum1966} and language understanding~\cite{winograd1972} began applying formal representations and inference techniques to human language~\cite{warren-pereira-1982-efficient}. This enabled computers to reason about the meaning of sentences, answer questions, and interact with users in more human-like ways, thus bridging the gap between symbolic AI and language understanding. These efforts opened the way in the development of semantic parsing~\cite{kamath2018survey}, a field dedicated to translating natural language into formal meaning representations.

The term \textit{Autoformalization} was originally introduced to describe the automatic formalization of mathematics with interactive theorem provers (proof assistants) such as Isabelle/HOL, Lean, Rocq, and Mizar, a field that has grown rapidly in recent years; there is increasing interest in employing deep learning to automate the task of formalization~\cite{wang2018first,szegedy2020promising,wu_autoformalization_2022,jiang2024multi}. Earlier, pioneering work employing techniques from machine learning had introduced the automation of the task of formalization referring to it as ``corpus-based translation between informal and formal mathematics''~\cite{kaliszyk2014developing} and 
``parsing between aligned corpora'', exploring the challenge: \textit{Is it possible to automatically parse informal mathematical texts into formal ones and formally verify them?} \cite{kaliszyk2015learning}. Autoformalization is closely related to semantic parsing~\cite{kaliszyk2017automating} and can be seen as an instance of semantic parsing. The distinctive feature of autoformalization would be that---unlike semantic parsing which encompasses a broad range of target formalisms including general purpose programming languages---it is restricted to target formal languages used in logical inference and automated reasoning. Independently of this distinction, with the development of deep learning approaches to natural language processing—particularly the emergence of transformer architectures—the task of translating informal language into formal representations is most often performed using large language models (LLMs)\footnote{The commonly used models include \texttt{Claude}, \texttt{Codex}, \texttt{DeepSeek}, \texttt{GPT3.5}, \texttt{GPT4}, \texttt{Llemma}, \texttt{Minerva}, \texttt{Mistral}, and \texttt{Mixtral}.}. This approach is more general and flexible, and less tied to a specific target language or domain than traditional semantic parsing methods; however, it comes at the cost of introducing new challenges in verifying their less restricted outputs.

Despite its growing use, current approaches to autoformalization remain largely ad hoc, and there is no---outside the original sense of formalizing mathematics---widely agreed-upon definition of what the term encompasses. This lack of clarity hinders communication across subfields of formal methods, proof engineering, and AI. Different communities may use the term to refer to subtly different tasks, or not explicitly consider their work to constitute autoformalization, making it difficult to compare methods, evaluate results, and transfer knowledge. Establishing a clear, comprehensive, and widely accepted definition would facilitate interdisciplinary collaboration and accelerate progress by enabling more systematic benchmarking, evaluation, and knowledge sharing. To address these issues, we propose a common framework for understanding autoformalization.

An additional motivation is the potential role of autoformalization in advancing the next generation of AI systems. Since the emergence of LLMs, there has been growing interest in using them as general-purpose reasoners. However, despite their impressive versatility in natural language processing, LLMs---trained primarily as next-token predictors---hallucinate, producing outputs that are logically inconsistent or factually incorrect. In response, ``reasoning'' models have been developed to address these shortcomings by incorporating techniques such as chain-of-thought prompting and reinforcement learning. While these models show improved performance on certain benchmarks, they still fall short in reliably performing complex logical inference~\cite{shojaee2025illusion} and remain costly to train. Furthermore, reasoning in these models is often difficult or impossible to verify formally, raising concerns about their use in high-stakes applications~\cite{NGUYEN2025}. In contrast, autoformalization leverages a task that large language models naturally excel at—translation—by mapping informal language into formal representations. This offers a potentially more efficient and interpretable alternative, or at least a complementary approach, to building general-purpose LLM-based reasoners.  By streamlining the interaction between language models and formal reasoning tools, autoformalization could play a key role in enhancing robustness and accuracy in domains such as mathematics, software verification, legal reasoning, and scientific discovery—where logical reasoning is essential.

\section{Review of Existing Usage and Definitions}

To propose a unified cross-disciplinary framework, we surveyed 81 research papers focused on the automatic translation of natural language into formal languages used for logical inference and automated reasoning. While we aimed to cover all relevant work in this area, we do not claim that the review is exhaustive. Our objective was not a detailed review of the included papers, but rather to examine both formal and informal definitions of \textit{autoformalization} and to identify tasks that can be interpreted as instances of autoformalization, even if the term itself was not explicitly employed. 

We included papers on \textit{autoformalization} as the term has been used since 2018, i.e., referring to the automatic translation of informal mathematics into ``formal and verifiable mathematical language''~\cite{wang2018first}. Furthermore, noting that the term \textit{autoformalization} has begun to gain traction outside the context of mathematical formalization~\cite{li2023logiclm}, we also included work on the translation of natural language into other languages that support formal verification, reasoning, or knowledge representation—such as Prolog, PDDL, or OWL—even if these are not languages of interactive theorem provers.

We categorized the reviewed papers into four broad subfields: formalization of mathematics using interactive theorem provers, logical inference and declarative programming, planning, and knowledge representation. These subfields were selected to reflect distinct research communities, each of which may conceptualize autoformalization differently and may have limited interaction or knowledge exchange with others. Nonetheless, the boundaries between these subfields are not always clear-cut. On the one hand, there is sometimes overlap between the categories (e.g., logical reasoning can be used for knowledge representation); on the other hand, additional categories may be warranted as the field evolves.

\paragraph{Formalization of Mathematics with Interactive Theorem Provers}

The term \textit{autoformalization} originated in the context of formalizing mathematics in 2018~\cite{wang2018first}, and as a result, it is widely adopted within this domain. Since the recent emergence of LLMs, users of proof assistants have been experimenting with using LLMs to assist with the formalization of mathematics, yielding mixed but potentially promising results
\cite{karatarakisleveraging}. All 32 papers reviewed for a definition from this subfield use the term \textit{autoformalization}. Most of works provide only informal definitions of autoformalization, which may reflect the fact that formalization of mathematics is intuitively well understood within the community and does not require defining. These informal definitions typically describe autoformalization as the task of automatically translating mathematical content from informal language into formal languages~\cite{wu_autoformalization_2022,jiang2022draft,agrawal_towards_2022,li_autoformalize_2024,zhou_dont_2024,ying_lean_2024,yang2024formal,liu2025safe,lu_process-driven_2024,poiroux_improving_2024,chan_lean-ing_2025,jiang_language_2025,azerbayev_proofnet_2023,wang2020exploration,yu2025formalmathbenchmarkingformalmathematical,bavarian2024typecheck,gontier2024stepproof,wang2025kimina}. Some authors also explicitly include machine-verifiability of the output as part of the definition~\cite{szegedy2020promising,murphy_autoformalizing_2024,liu_rethinking_2024,liu_atlas_2025,patel_new_2023,cunningham_towards_2023,chang2024rethinking,weng2025autoformalization,lu2024formalalign,jiang2024multi,gadgil2022towards}. Only one group of authors attempted a formal definition~\cite{zhang2024consistent,zhang_formalizing_2025}.


\paragraph{Logical Inference and Declarative Programming}


In this category, we reviewed 30 papers. The target formalisms include first order logic~\cite{li2023logiclm,lalwani_autoformalizing_2024,han_folio_2022,hahn_formal_2022,yang_coupling_2023,olausson_linc_2023,hupkes2024learning,thatikonda_strategies_2024,ryu_divide_2024,quan_verification_2024,brunello2025evaluating,lin2025code4logic,lee_safeguarding_2025}; declarative programming language, Prolog~\cite{mensfelt_autoformalization_2024,mensfelt_gama_2024,borazjanizadeh2024reliable} and answer set programming (ASP)~\cite{li2025logic,llasp2024,nl2asp2024,llm_asp_2023}. Another main line of work focuses on translating
problems from natural language into temporal logics
such as LTL~\cite{pnueli1977temporal} and CTL~\cite{clarke1986automatic}. These languages extend classical logic with temporal operators to reason about properties that hold over time, with LTL describing linear sequences of states and CTL describing branching time structures where multiple futures are possible from each state. There exist a significant number of recent works focused on translating natural language descriptions to these languages \cite{chen_nl2tl_2023,data_efficient_2023,guiding_llm_2024,liu_lang2ltl-2_2024,mavrogiannis_cook2ltl_2024,xu_learning_2024,safety_ltl_2025,soroco2025pde,wang2025let,wang_conformalnl2ltl_2025}; one the reasons of popularity of temporal logics is its relevance in industrial applications.

Only 8 of reviewed papers explicitly used the term \textit{autoformalization}, and all of them defined it informally—typically as the translation of informal or natural language into formal representations~\cite{li2023logiclm,lalwani_autoformalizing_2024,mensfelt_autoformalization_2024,mensfelt_gama_2024,soroco2025pde,olausson_linc_2023,lee_safeguarding_2025,quan_verification_2024}. Among the remaining works, one paper formally defines the task~\cite{liu_lang2ltl-2_2024}, while the rest describe it informally as translation from natural language into a specific formalism. Notably, even though these papers do not use the term \textit{autoformalization}, their definitions are consistent with those that do.

\paragraph{Planning}

The Planning Domain Definition Language (PDDL) is the de facto standard formalism within the AI planning community \cite{PDDL}, encompassing various extensions ranging from basic STRIPS to first-order, probabilistic, and temporal logics \cite{PDDLFox}. 

A PDDL planning instance consists of a domain file and a problem file. The domain file specifies predicates (facts) and actions applicable across the entire domain. Actions are defined through preconditions and effects, encapsulating a labeled transition system that alters the truth values of a set of facts. The problem file describes the particular instance, specifying the initial state and the goal conditions.

PDDL formally characterizes logical planning problems where the task is to generate a sequence of actions (a \emph{plan}) to transition from the initial to a goal state. Importantly, PDDL is purely descriptive; it does not inherently provide solutions. Therefore, planners like \cite{helmert2006fast} are required to compute plans. This separation allows LLMs to effectively tackle complex combinatorial problems by translating them into PDDL representations, leveraging specialized logic solvers and thus combining the strengths of LLMs (natural language translation) and automated planning systems (combinatorial reasoning).

In the planning category, we included 16 papers. None of them use the term ``autoformalization''. One of the works provides a formal definition of the task~\cite{hu_text2world_2025} as defining a mapping function generating a world model from a natural language description. The rest of the works define the task informally~\cite{cllamp2024extract,liu_llmp_2023,aghzal_survey_2025,smirnov_generating_2024,xie_translating_2023,mahdavi_leveraging_2024,huang_limit_2024,hu_text2world_2025,lin_text2motion_2023,oswald_large_2024,silver_generalized_2024,guan2023leveraging,sikes_creating_2025,de_la_rosa_trip-pal_2024}. In some cases the task may encompass generating both domain and problem files~\cite{smirnov_generating_2024}, while in others only PDDL snippets are being instantiated, e.g., goals in ~\cite{xie_translating_2023}. A related but separate line of research involves using LLMs as planners themselves, generating plans from PDDL inputs. This direction falls outside the scope of the present work.

\paragraph{Knowledge Representation}

Several works that we surveyed~\cite{groza_ontology_2023,doumanas_sar_2024,caufield_structured_2024,saaedizade_navigating_2024} focus on translating natural language descriptions of real-world domains into ontologies using the Web Ontology Language (OWL)~\cite{w3c:owl2-overview}, 
a formal language used to define and share structured, machine-readable knowledge in terms of individuals,  concepts, and relationships between them.
Some other works produce ontologies expressed in RDF or RDFS~\cite{brickley2014rdfs}.
RDF allows us to express Knowledge Graphs, where nodes represent named entities, node labels represent their individual properties (called ``types'' or ``classes''), and labelled edges represent relations between them. RDFS enriches RDF with information about subsumption, domain, and range relations between classes. Both RDF and RDFS can be seen as subsets of OWL. Works such as \cite{ells_commonsense_2024,sadeq_leveraging_2025} translate natural language to RDFS and RDF ontologies, respectively.
Others express ontologies in ad-hoc languages \cite{tang_domain_2023,lippolis_ontology_2025}, which seem to be expressible in OWL or RDFS. In all these works, syntactic validity is checked automatically, and semantic is checked manually, or by running certain conjunctive queries on the produced ontology. None of the reviewed works in this category employs the term \textit{autoformalization} to denote the translation task.


\paragraph{Summary}

As observed, definitions across domains exhibit notable similarities, even when the task is not explicitly labelled as \textit{autoformalization}. At the core, these tasks involve the automatic translation of natural language into a target language that supports logical inference and automated reasoning. Building on these shared foundations, we now propose a unified conceptual framework that can encompass this broad class of tasks.

\section{Proposed Definition}

To define autoformalization, we begin by first establishing a definition of formalization. Then, we demonstrate how selected case studies---simple examples from various domains ---align with this definition.

\subsection{Definition of Autoformalization}

We introduce a general definition intended to capture the essential components of autoformalization across a broad range of domains and settings, while abstracting away from implementation-specific or domain-specific details.

\begin{definition}
\emph{Formalization} from informal language $L_i$ to formal reasoning language $L_f$ with respect to a semantic equivalence criterion $E$ is the transformation of an expression in a domain-specific subset of $L_i$ into a well-formed and valid expression in $L_f$ that is semantically equivalent according to $E$.
\emph{Autoformalization} is formalization performed automatically by a computing system.
\end{definition}

Our definition involves four parameters: the informal language $L_i$, a domain-specific subset of $L_i$,
the formal language $L_f$, and the semantic equivalence criterion $E$. These are treated as primitives
without formal definition. However, we proceed to clarify their meanings, provide examples,
and explain our rationale for including them in the definition.

\paragraph{Informal language}
An informal language $L_i$ is a collection of meaningful expressions, such as the set of all grammatically well-formed and semantically coherent texts in English. This abstraction deliberately ignores more fine-grained aspects,
such as lexical categories or compositional syntax. Membership in informal languages is generally neither
decidable nor computable, and their boundaries may be vague or context-dependent. Sometimes we consider 
\emph{semi-formal} languages: informal languages (like plain English) mixed with
elements of a formal language. For example, the mathematical language of proofs is typically semi-formal.

Our definitions refers to a domain-specific subset of $L_i$ because, 
in a given autoformalization setting, we typically do not seek to formalize all possible expressions in $L_i$,
but only a specific subset of interest that share a common conceptual framework or subject matter.
For example, within plain English, we may consider: 
descriptions of two-player games, specifications of procedural planning settings, and
natural language descriptions of real-world systems in terms of their objects, properties,
and relations (i.e.\ informal ontologies), among many other examples.

\paragraph{Formal reasoning language}
The formal reasoning language $L_f$ is an enumerable set of expressions
typically specified by a grammar or formation rules, and is accompanied by a formal
semantics that assigns precise and unambiguous meaning to each well-formed expression.
It is also associated with a reasoning apparatus---usually a set of inference
or transformation rules---that enables the derivation of new expressions from
existing ones while preserving semantic properties. 
Examples of \( L_f \) include:
\begin{itemize}
  \item languages of proof assistants, i.e., interactive theorem provers (e.g., Lean, Isabelle, Mizar, Rocq);
  \item fully formal logics (e.g., propositional logic, FOL, LTL);
  \item logic programming and declarative languages (e.g., Answer Set Programming, Prolog);
   \item planning and CSP languages (e.g., PDDL, MiniZinc)
  \item knowledge representation languages and formalisms (e.g., OWL, RDF, Situation Calculus, Event Calculus).
\end{itemize}
The choice of formal language $L_f$ in a given autoformalization setting is
guided by two considerations: first, it must be capable of representing all relevant
information in the domain-specific subset of $L_i$; second, it must support
reasoning in a way that allows properties of interest to be verified, both in
terms of computational complexity and availability of appropriate verification tools.
For example, if our goal is to formalise descriptions of planning scenarios, $L_f$ might be
a planning modeling language such as STRIPS.

\paragraph{Semantic equivalence criterion}
The semantic equivalence criterion $E$ specifies what it means for the obtained formal expression
from $L_f$ to preserve the intended meaning of the informal input from $L_i$.
While the goal is for both expressions to `mean the same thing,' formal languages
are usually less expressive than informal languages and cannot capture every aspect
of the original meaning. Hence, in practice, $E$ is designed to ensure that the
formalization preserves the aspects of meaning that are most relevant for the task at hand.
For example, in an interactive theorem prover, $E$ may require that the
formalization contains sufficient detail to ensure that formalisations of
correct proofs (expressed informally in $L_i$) can be verified by a given tool or
reasoning algorithm. 

Specifying the semantic equivalence criterion can be challenging, as it requires
defining conditions related to the meaning of informal language, which is inherently
ambiguous. To address this, we introduce a related concept, the \emph{validation criterion} $V$,
which serves as a computable approximation of $E$. This distinction is important: $E$
defines what it means for an informal and formal expression to ``mean the same thing'' 
as a formal expression, whereas $V$ provides a practical means of verifying whether a
given pair of informal and formal expressions satisfies this equivalence.

We conclude this section with several observations that clarify and extend the scope of the definition.

\begin{remark}
An informal expression in $L_i$ may be formalized in two or more distinct ways into $L_f$ which may differ in their choice of names, syntactic structure, or level of abstraction. Conversely, two distinct informal expressions may be formalized into the same expression in $L_f$; this may happen, for example, if they describe the same state of affairs according to the semantic equivalence criterion $E$, reflecting the inherent variability and redundancy of natural language.
\end{remark}

\begin{remark}
Systems for autoformalization need not be deterministic: the same informal input may be mapped to different formal outputs in different runs. These outputs may or may not be equivalent to one another, but each must satisfy the equivalence criterion $E$ with respect to the input. In practice, it is often desirable for a system to prefer outputs that are syntactically or structurally minimal, promoting conciseness and efficiency.
\end{remark}

\begin{remark}
Although our definition is concerned with equivalence between complete informal and formal expressions, many applications require equivalence to hold at a finer granularity. For example, a multi-sentence informal description of a mathematical argument may be formalized as a sequence of formal statements, each corresponding to a specific sentence or sub-argument. Such fine-grained correspondence should be subsumed within the semantic equivalence criterion $E$.
\end{remark}

\subsection{Case Studies}

To further illustrate  the flexibility of our definition and its applicability across a range of settings, we next show how it can be applied to capture existing examples in the literature. The case study analysis will focus on what constitutes informal input, formal target language, and how semantic equivalence is evaluated, while abstracting away from implementation details such as the application of in-context learning and self-debugging feedback loops.  

\subsubsection{Formalization of Mathematics with Interactive Theorem Provers}

The following example comes from a seminal paper on autoformalization~\cite{wu_autoformalization_2022}.
In this example, the informal language $L_i$ is the natural mathematical language (plain English extended with mathematical symbols) and the relevant subset can be considered as a chosen set of mathematical problems (the example at hand coming from the 1987 International Mathematical Olympiad),
with the target language $L_f$ in this case being that of Isabelle/HOL \cite{nipkow2002isabelle}. Isabelle/HOL is the 
 most widely used instance of Isabelle, an interactive theorem prover for formal verification and proof construction. Isabelle/HOL provides a higher-order logic theorem proving environment. In this context, $L_f$ is Isabelle's structured proof language \textit{Isar} that includes symbols such as keywords, mathematical symbols, and logical operators, and has a formally defined syntax.  The reasoning apparatus is realised by various automatic tools incorporated in Isabelle/HOL
 (and by a number of external automatic theorem provers and satisfiability-modulo-theories solvers which can be called by the Sledgehammer tool featured by Isabelle/HOL \cite{blanchettesledgehammer}).

In our example \cite{wu_autoformalization_2022}, the input expression to be formalized in $L_i$ was as follows:

\begin{italicquote}
Prove that there is no function \( f \) from the set of non-negative integers into itself such that \( f(f(n)) = n + 1987 \) for every \( n \).
\end{italicquote}

The corresponding formalization in Isabelle/HOL produced by the Codex LLM tool was as follows:

\begin{quote}
\small
\begin{verbatim}
theorem
fixes f :: "nat \<Rightarrow> nat"
assumes "\<forall> n. f (f n) = n + 1987"
shows False
\end{verbatim}
\end{quote}

The statement \texttt{shows False} indicates that the assumption of the described function's existence leads to a contradiction.

The semantic equivalence criterion $E$ in this case requires that all important mathematical details from the original theorem statement are captured in the formal statement. This is crucial because when we later formalize a proof and verify its validity, all essential claims that the proof must establish need to be correctly represented in the theorem statement.

Two validation criteria $V$ were used here to evaluate semantic correctness. First, BLEU scores~\cite{papineni2002bleu} were used, which measure surface-level similarity between the autoformalized statement and the ground truth (i.e.\ a manual formalization deemed correct by human experts), but not direct semantic equivalence. The second validation criterion was manual evaluation of selected examples. While manual evaluation reliably assesses semantic equivalence, it is costly and does not scale well. This highlights a fundamental challenge in autoformalization: automating the evaluation of semantic
correspondence.

\subsubsection{First-Order Logic}

In~\cite{li2023logiclm} authors explicitly state 
that their goal is to extend autoformalization into logical reasoning. In the example from the paper, their relevant subset of natural language ($L_i$) consists of questions about real-word objects and shapes.
The target formal system $L_f$ is \textit{Prover9}, which is based on first-order logic syntax and a special signature of relevant predicates  (e.g., \texttt{Square}, \texttt{FourSided}, \texttt{Shape}). 
The reasoning apparatus for this language is instantiated by Prover9; it is a sound calculus capable of verifying logical entailments within this formal system.

Consider the following example from the FOLIO benchmark:

\begin{italicquote}
\textbf{Context:} All squares have four sides. All four-sided things are shapes.\\
\textbf{Question:} Based on the above information, is the following statement true, false, or uncertain? \textit{All squares are shapes.}\\
\textbf{Options:} A) True \quad B) False \quad C) Uncertain
\end{italicquote}

This informal input is translated into the following expression, consisting of three different sets of first-order logic formulas:

\begin{quote}
\footnotesize
\textbf{Predicates:}\\
\texttt{Square(x)} — \( x \) is a square\\
\texttt{FourSided(x)} — \( x \) has four sides\\
\texttt{Shape(x)} — \( x \) is a shape\\

\textbf{Premises:}\\
\( \forall x (\texttt{Square(x)} \rightarrow \texttt{FourSided(x)}) \)\\
\( \forall x (\texttt{FourSided(x)} \rightarrow \texttt{Shape(x)}) \)\\

\textbf{Conclusion:}\\
\( \forall x (\texttt{Square(x)} \rightarrow \texttt{Shape(x)}) \)
\end{quote}

Prover9 then verifies that the conclusion logically follows from the premises via transitive implication. Hence, the system formally determines that the correct answer is \textbf{True (A)}. In the paper, the semantic equivalence criterion $E$ relies on all relevant properties of the objects described in the question being captured by the logical formulas, but wihtout introducing spurious properties. 
The validation criterion $V$ used to measure semantic equivalence consisted in comparing the answer derived from reasoning on the formalized problem to the ground truth correct answer. This approach enables automated evaluation. However, reaching the correct answer does not necessarily guarantee semantic equivalence.

\subsubsection{Logic Programming} The next example involves formalizing strategic interactions that can be modelled as bimatrix games in order to reason about them formally in Prolog~\cite{mensfelt_gama_2024}. In this case, $L_f$ is the set of valid Prolog programs. The semantics of Prolog is well-defined and the reasoning apparatus relies on logical inference; in this case, it is realised by the SWI-Prolog solver.

In the following example, the relevant set of expressions involves description of game-like interactions, expressed in natural language ($L_i$). The specific example involves the classical Battle of the Sexes game:

\begin{italicquote}
A couple is deciding how to spend their evening together. One prefers to go to the opera and the other prefers to go to a football game, but both prefer to be together rather than apart. If they both choose the opera, the opera lover gets a payoff of 2 units of happiness, and the football lover gets 1 unit. If they both choose the football game, the football lover gets 2 units of happiness, and the opera lover gets 1 unit. If they choose different events, they both get 0 units of happiness.
\end{italicquote}

The formalization of this expression results produces a set of domain-dependent predicates that, together with domain-independent predicates, allow for reasoning about the interaction. The snippet below, which is part of the formalization output, shows the payoff matrix for the game and the subset of predicates defining the initial state, legal moves and the effects of moves: 
\begin{lstlisting}
...
payoffBOS(opera, opera, 2, 1).
payoffBOS(football, football, 1, 2).
payoffBOS(opera, football, 0, 0).
payoffBOS(football, opera, 0, 0).

initial(s0).

initially(player(p1), s0).
initially(player(p2), s0).
initially(role(p1, opera_lover), s0).
initially(role(p2, football_lover), s0).
...
initial(s0).
...
legal(select(P, opera), S):-
    holds(player(P), S),
    holds(control(P), S).
...
effect(did(P, M), select(P, M), S).
...
abnormal(control(P), select(P, M), S).
...
\end{lstlisting}
The semantic equivalence criterion refers to the correct formalization of game rules and parameters (i.e., payoffs), such that player 1 selecting opera and player 2 selecting football should result in payoffs of (0,0) when we query the generated formalization. The verification $V$ of the semantic equivalence criterion is automated by querying all possible combinations of actions and comparing the resulting payoffs to ground-truth payoffs for each pair of actions.

This method of evaluation is both automated and guarantees practical semantic equivalence. However, it applies only to a narrow class of problems—bimatrix games—and requires manually implementing a validation procedure for each class of games.

\subsubsection{Planning}

For planning we chose as an example the one used in ~\cite{oswald_large_2024}.  
In this case, the formal language $L_f$ consists of PDDL actions. PDDL is a formal language for representing classical planning problems, where states are described using first-order logic atoms, actions define state transitions, and the goal is to find a sequence of actions that transforms the initial state into one satisfying goal conditions. The syntactic framework of PDDL defines the structure of predicates, object types, and action definitions with preconditions, and effects. For actions, the syntax consists of `(:action $<$action-name$>$ ...)', where action parameters are defined, followed by the `:precondition' (the conditions that must hold before the action can be executed) and the `:effect' (the changes that result from executing the action). The formal reasoning apparatus for PDDL included the planners that interpret the generated PDDL domain and solve planning problems. Specifically, the planners used in the paper include the K* planner~\cite{lee2023k} (part of the semantic verification checks the correctness of 100 alternative plans for each problem instance) for plan generation.

The example informal language input $(L_i)$ is mostly natural language, extended with symbols that represent names that must appear in the corresponding formal expression. Note that this is aided by adding a list of allowed predicates to the informal action description prompt:

\begin{italicquote}
Allowed Predicates: 

(in-city ?loc - place ?city - city):a place loc is in a city.

(at ?obj - physobj ?loc - place) : a physical object obj is at a place loc.

(in ?pkg - package ?veh - vehicle) : a package pkg is in a vehicle veh.

The action, ``FLY-AIRPLANE'' will fly an airplane from one airport to another. After the action, the airplane will be in the new location.
\end{italicquote}

The output is a PDDL representation of the described action, where the given action's preconditions and effects are formalized to model the planning domain:

\begin{lstlisting}[style=pddl] % Apply the pddl style defined earlier

(:action FLY-AIRPLANE 
  :parameters (?airplane - airplane 
  ?loc-from - airport ?loc-to - airport)
  :precondition (at ?airplane ?loc-from)
  :effect (and(not(at ?airplane ?loc-from))  (at ?airplane ?loc-to)) )
\end{lstlisting}

As with logical entailment in first-order logic, planning also requires a notion of semantic equivalence. However, domain equivalence in planning involves action behaviour as well as purely logical consequence. To evaluate approximate semantic alignment, the authors use a \textit{heuristic domain equivalence} test. For a reconstructed domain \( \mathbf{D_o}' \) generated by replacing an action \( a \) in the original domain \( \mathbf{D_o} \), plan-based reasoning is used to determine equivalence:

\begin{itemize}
    \item Let \( \Pi \) be a planning problem solvable under \( \mathbf{D_o} \), and \( \Pi' \) its analog using \( \mathbf{D_o}' \). Let \( P \) and \( P' \) be subsets of valid plans generated from \( \Pi \) and \( \Pi' \), respectively.
    \item Each plan \( \pi \in P \) is checked for validity under \( \Pi' \), and each \( \pi' \in P' \) is validated under \( \Pi \), using the VAL tool~\cite{howey2004val}.
    \item If all cross-validations succeed, then \( \text{Semantics}(L'_f) \approx \text{Semantics}(L_f) \), and the output action is deemed to be semantically equivalent in terms of planning behaviour.
\end{itemize}

This procedure defines a relaxed but practical notion of equivalence suitable for evaluating model outputs in symbolic planning systems. As with logical reasoning, selecting a formally well-structured action that ``looks right'' is not sufficient: only through validation methods like this can the semantic faithfulness of the model's output be determined.

\subsubsection{Knowledge Representation}

One approach to formalizing domain knowledge involves translating natural language (NL) statements into OWL axioms. \cite{groza_ontology_2023} demonstrate how fine-tuned large language models can facilitate this translation by converting concise declarative sentences into OWL axioms---expressed in \emph{Functional Syntax}, one of several equivalent styles to write OWL axioms. Their system uses a GPT-3 model trained on a dataset of 150 prompt–response pairs. The training data covers various ontology constructs, including class declarations, subclass hierarchies, object properties, cardinality restrictions, and disjointness.

The Web Ontology Language (OWL) is a formalism used to model knowledge about a domain in terms of \emph{named individuals}, their individual properties (called \emph{concept names} or \emph{classes}), and the binary relationships between them (called \emph{roles} or \emph{object properties}). OWL includes logical constructors that allow the definition of complex concepts and role expressions from simpler ones. 
The formal language $L_f$ in this setting is the set of OWL ontologies:  a set of logical assertions, typically divided into three categories: \emph{TBox assertions}, which express general class relationships such as subclassing or equivalence; \emph{ABox assertions}, which describe facts about individual instances; and \emph{RBox assertions}, which define properties of roles, such as transitivity or symmetry. They often include also optional declaration statements (which can be inferred). Key reasoning tasks include consistency checking   (determining whether the ontology admits a model) and classification (computing the taxonomy of concept names).

In this setting, the semantic equivalence criterion requires that all information described in the informal description is correctly captured by the ontology. The verification is often established informally by ontology engineers, who manually verify that the ontologies capture all relevant knowledge. However, reasoning tools play a crucial role in this process by checking for consistency and ensuring that the ontology entails all expected consequences while avoiding unintended ones. Ontology development is often an iterative process, alternating addition of axioms and reasoning to assess their impact.

To illustrate the approach of \cite{groza_ontology_2023}, we recapitulate the output
of their system on the sentence \emph{Anna and Lana are each other’s sisters}:
{ \footnotesize
 \begin{align*}
 \texttt{Declaration(ObjectProperty(:has\_sister))}\\
 \texttt{Declaration(NamedIndividual(:Anna))}\\
 \texttt{Declaration(NamedIndividual(:Lana))}\\
 \texttt{ObjPropAssert(:has\_sister :Anna :Lana)}\\
 \texttt{ObjPropAssert(:has\_sister :Lana :Anna)}
 \end{align*}
}
The first statement identifies the role (binary predicate) \texttt{has\_sister} and
named individuals \texttt{Anna} and \texttt{Lana}. Then it generates two additional 
assertions to represent the symmetry implied in the sentence.

In this setting, the semantic equivalence criterion $E$ requires that all relational
information in the informal description be correctly captured in the OWL ontology.
In practice, this cannot be formally measured, so the paper establishes a validation
criterion $V$ based on ontology engineers manually verifying that the ontology encodes
all relevant knowledge. In general, however, such manual $V$ can be supplemented with
an automated component: assuming the informal knowledge is consistent---as is usually
the case---automated reasoning tools can check consistency of the formalization.
Furthermore, when the expected consequences and undesirable outcomes of the knowledge base
are known and have admissible formalizations, ontology reasoners such as Hermit~\cite{hermit} can be used to verify formally
that the ontology both entails all intended consequences and avoiding unintended ones.

\section{Challenges and Open Problems}

While there is a growing body of work aimed at advancing autoformalization~\cite{patel_new_2023,zhang2024consistent,li_autoformalize_2024,tarrach2024more,chan_lean-ing_2025,jiang2024multi,lu2024formalalign,poiroux_improving_2024,chang2024rethinking}, several fundamental challenges still offer exciting research opportunities that will shape the field's future development. We next provide a selective overview of these challenges and opportunities.

\subsubsection{Semantic Verification}
The key challenge in autoformalization remains verifying equivalence of the informal language and the formal representation. Current approaches rely mostly on human supervision, either through direct assessment or comparison with reference solutions; this is sometimes enhanced with automated checks such as consistency checking of the extracted formalism. Fully formal and automated verification of semantic equivalence, however, appears impossible even in principle, since natural language and its intended meaning are inherently informal.

An open question is whether it is possible to train intelligent systems to acquire the dual competence that humans possess: understanding both formal and informal domains, and being able to assess correspondence between them. Such verification capabilities are in principle distinct from the current practice of autoformalization, but they could be integrated directly into the autoformalization process rather than applied post-hoc.
Although formally unverifiable, using meta-level reasoning to relate formal and informal representations could significantly reduce supervision and move us closer to automated human-level semantic verification.

\subsubsection{Target Formal Languages}

Another challenge in autoformalization is selecting an appropriate target formal language for the required task. This choice depends critically on the intended use case: who will interact with the formalized text, and what computational tasks will be performed with it? For instance, if the goal is interactive theorem proving, the availability and sophistication of verification tools as well as the contents of formal proof libraries become critical. On the other hand, if human mathematicians will work with the output, considerations of readability and established community conventions become essential.

The selection process must also balance expressiveness against computational tractability. The target language must be sufficiently expressive to capture all the relevant semantic content of the natural language text.
However, there exists a fundamental tension between expressive power and computational complexity of verification, potentially making automated reasoning and verification very difficult or even undecidable.
Given these constraints, an open problem is exploring the development of new formal languages designed specifically for autoformalization, which might offer better balance points in the expressiveness-tractability spectrum.

\subsubsection{Scalability and Integration with Mathematical Practice}

Current autoformalization systems work well on curated benchmarks but often fail to scale to real-world practice. For example, in the context of mathematical proof, the literature is dominated by incomplete proofs, informal reasoning steps, and proof sketches, and so their formalization requires contextual understanding and gap-filling, capabilities that current systems struggle with, particularly in specialized mathematical fields.

This scalability gap reflects a deeper uncertainty about how autoformalization tools should integrate into research workflows. Without clarity on whether autoformalization should assist proof development, literature review, verification, hypothesis testing, or enhancing the reliability of AI systems, we risk building systems that are technical achievements but remain peripheral to the work of intended users.

\subsubsection{Interactive Formalization}
Many of the current autoformalization approaches treat the task as a one-shot translation problem, attempting to convert natural language text directly into formal representations. As we have already discussed, this  paradigm struggles with the inherent ambiguity and context-dependence of natural or mathematical language. A more promising direction---and an approach to the problem of semantic verification---involves systems that can engage in dialogue with users, asking clarifying questions about ambiguous terms, implicit assumptions, or intended interpretations.

Such interactive systems would require significant advances in both natural language understanding and theorem prover interfaces. The system should recognise when clarification is needed and formulate meaningful questions to the user, which then would be incorporated to iteratively refine the formalization. 
Early feedback would also prevent the system from pursuing incorrect interpretations too far before correction.

\subsubsection{Cross-Domain Transfer and Generalization}

As we have shown in this paper, current autoformalization approaches are often domain-specific, with systems designed separately for areas such as mathematical reasoning, planning, logic, etc. A major challenge is developing methods that can transfer reasoning patterns and formalization strategies across these different areas. Our present analysis of the notion of autoformalization shows that the underlying formal structures and reasoning patterns in these areas often share commonalities that could be exploited for more general autoformalization systems.

This, however, is still a big open challenge that requires a more systematic understanding of the structural similarities in the way autoformalization is used across fields. Success in this area would enable more robust and widely applicable autoformalization tools that could adapt to new domains with minimal additional training. We envision our current work, which establishes a common framework for the definition of autoformalization, as a first step in that direction.

\section{Conclusions}

In practice, we judge successful formalization by its fruits: Does it help us solve problems? Does it support verification? Does it reveal new connections? Does it eliminate disputes? Does it enable computation? 
Formalization is less about focusing on mere translation and more about creating tools that extend our and AI systems' reasoning capabilities in useful ways.

One of the most promising applications is enhancing the capabilities of LLMs, developing robust general-purpose reasoners, and, as one of the consequences, enhancing the safety of agentic AI. However, achieving this through autoformalization presents many challenges. A common framework could accelerate progress by facilitating knowledge exchange between different fields and enabling transfer learning.

As we have demonstrated, translating problems into different formalisms for logical inference and automated reasoning shares similarities and common challenges. Based on the reviewed papers, we propose a preliminary definition that encompasses various types of formalisms. This definition is not intended to be definitive; rather, our aim is to initiate a cross-disciplinary dialogue on how best to conceptualize autoformalization, articulate its desired properties, and promote effective interdisciplinary collaboration.

\section*{Acknowledgments}

This work was supported by a Leverhulme Trust International Professorship Grant (LIP-2022-001).

\bibliographystyle{kr}
\bibliography{references}

\end{document}